\documentclass[a4paper,conference]{IEEEtran}
\usepackage{cite}

\ifCLASSINFOpdf
\else
\fi

\usepackage{times}
\usepackage{epsfig}
\usepackage{graphicx}
\usepackage{float}
\usepackage{amsmath}
\usepackage{amssymb}
\usepackage{multicol}
\usepackage{siunitx,booktabs}
\usepackage{subcaption}
\usepackage{svg}
\usepackage{enumitem}
\usepackage[pagebackref=true,breaklinks=true,letterpaper=true,colorlinks,bookmarks=false]{hyperref}

\begin{document}
%
\title{Towards Real-Time Analysis of Broadcast Badminton Videos}

\author{\IEEEauthorblockN{Nitin Nilesh}
\IEEEauthorblockA{CVIT, KCIS\\
IIIT Hyderabad\\
nitin.nilesh@research.iiit.ac.in}
\and
\IEEEauthorblockN{Tushar Sharma}
\IEEEauthorblockA{Product Labs\\
IIIT Hyderabad\\
tushar.sharma@iiit.ac.in}
\and
\IEEEauthorblockN{Anurag Ghosh}
\IEEEauthorblockA{CVIT, KCIS\\
IIIT Hyderabad\\
anurag.ghosh@research.iiit.ac.in}
\and
\IEEEauthorblockN{C. V. Jawahar}
\IEEEauthorblockA{CVIT, KCIS\\
IIIT Hyderabad\\
jawahar@iiit.ac.in}}


%


\maketitle

\begin{abstract}
Analysis of player movements is a crucial subset of sports analysis. Existing player movement analysis methods use recorded videos after the match is over. In this work, we propose an end-to-end framework for player movement analysis for badminton matches on live broadcast match videos. We only use the visual inputs from the match and, unlike other approaches which use multi-modal sensor data, our approach uses only visual cues. We propose a method to calculate the on-court distance covered by both the players from the video feed of a live broadcast badminton match. To perform this analysis, we focus on the gameplay by removing replays and other redundant parts of the broadcast match. We then perform player tracking to identify and track the movements of both players in each frame. Finally, we calculate the distance covered by each player and the average speed with which they move on the court. We further show a heatmap of the areas covered by the player on the court which is useful for analyzing the gameplay of the player. \par

Our proposed framework was successfully used to analyze live broadcast matches in real-time during the Premier Badminton League 2019 ({\bf PBL 2019}), with commentators and broadcasters appreciating the utility.
\end{abstract}


%
\IEEEpeerreviewmaketitle

\begin{figure}
    \centering
    \includegraphics[width=0.45\textwidth, height=0.40\textwidth]{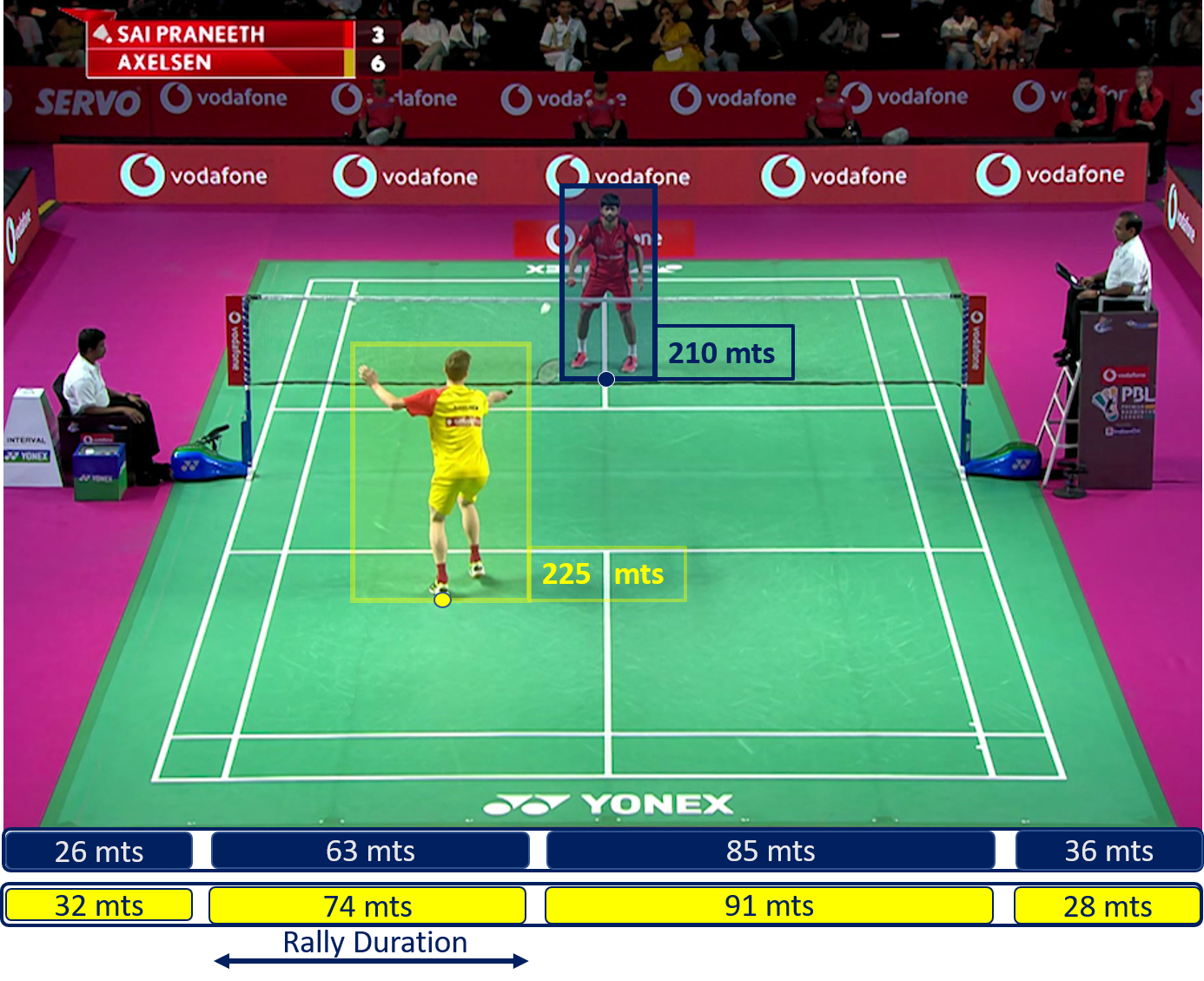}
    \caption{An overview of the distance travelled by the player from the start of a set. Player in blue bounding box refers to the ``top-player" and player in yellow bounding box refers to the ``bottom-player". Dot at the center-bottom of the player's bounding box represents their current location. The ``Rally Duration" is the time length of a particular rally. Four such rallies are depicted for reference. \textit{(best viewed in color)}}
    \label{fig:intro}
    \vspace{-3mm}
\end{figure}

\section{Introduction}
Sports analysis is one of the most challenging tasks in computer vision and several different approaches to sports analysis have been proposed. Sport analyses include match summarization\cite{tjondronegoro2004highlights}, event prediction \cite{giancola2018soccernet}, ball tracking\cite{maksai2016players}, structured analysis of the game\cite{ghosh2018towards} etc. This kind of analysis has proven to be useful in areas such as training and coaching,\footnote{\href{https://www.newscientist.com/article/dn28048-ai-football-manager-knows-how-different-teams-play-the-game/}{https://www.newscientist.com/ai-football}} and has also been able to provide help to the referee during the game. Apart from the types of analysis mentioned above, research topics like analysis and prediction of how groups of players move \cite{felsen2017will}, automatic identification of key stages \cite{sharma2017automatic} in a game for gathering statistics or automating the control of broadcast cameras \cite{gandhi2010real} etc. have also been explored. Game statistics are important to both viewers as well as the players and their coaches. For viewers, having access to accurate statistics improves viewing experience and helps them understand the more technical aspects of the sport like complex strategies and playing styles. It also aids the players and their coaches to formulate better game strategies and improve performance. Analysing an opponent's gameplay also helps players to adapt their style of play to suit the particular opponent. In badminton, games statistics include number of wins, unforced errors, successful smashes, distance covered, speed of the serve and so on. \par

While considerable research has been published regarding badminton analysis, the one common limitation of all existing methods is that they are applied post-factum on previously recorded match videos. This prevents the use of these methods to show statistics during a live broadcast match. The ability to compute these statistics in real-time from a live broadcast match not only makes the game more exciting for viewers but also allows the players to dynamically modify their strategies mid-game. However, broadcast videos are captured from multiple viewpoints for the best viewing experience and their `unstructured' nature along with the rapid movement and complex human pose and motion makes real-time analysis a complex task. Moreover, broadcast delay for a typical badminton match is less than 15 seconds, usually 7 seconds \footnote{\href{https://en.wikipedia.org/wiki/Broadcast_delay}{https://en.wikipedia.org/wiki/Broadcast\_delay}}. Many of the existing analysis methods are constrained by these challenges. Our main focus in this work is to provide real-time analysis framework for the broadcast videos which is expected to generalize to almost all the badminton games captured from broadcast camera view.\par 

In this work, we introduce an approach which generates heatmaps of player movements, calculates distance run by players and the average speed of payer movements for the game of badminton (figure \ref{fig:intro}). We perform point segmentation, player detection and homography based techniques to compute the above defined metrics.\par

The major contribution of this paper is as follows:
\begin{enumerate}[itemsep=1pt]
    \item We propose an end-to-end framework (figure \ref{fig:approach}) to analyze live broadcast match videos, the components of which can be trained using pre-recorded badminton matches. Unlike previous approaches, our method uses visual cues only and does not rely on any hardware setup.
    
    \item Using recent advancements in object detection, we predict players' location, their heatmap across the court and distance covered by each player during gameplay. We do these analyses for both singles and doubles matches.
    
    \item We tested our framework in a real-time scenario and analyzed live broadcast matches that were conducted as a part of the Premier Badminton League 2019.
    
    \item We introduce two new datasets, one each for singles and doubles matches covering a wide range of broadcasting camera angles, court and lighting conditions, and featuring several different players. We also introduce a set of 25 rallies badminton match video of our own recording having the ground truth data for the distance covered by the players in each rally. 
\end{enumerate}

\section{Related Work}
\noindent \textbf{Sports analysis and Player Localization: }Traditionally, work in analyzing sports videos has focused on either tracking players\cite{shitrit2011tracking} or balls\cite{maksai2016players} to analyze game formations or skill level of individual players.
Among all sports, racket sports has received a lot of attention with applications in video summarization and highlight generation\cite{ghanem2012context}. Yoshikawa et al.\cite{yoshikawa2010automated} implemented serve scene detection for badminton games using specialized overhead camera setup. Chu et al.\cite{chu2017badminton} performed semi-automatic badminton video analysis by court and player detection, and clustering player strategy into offensive or defensive by classifying strokes. Mlakar et al.\cite{mlakar2017analyzing} performed shot classification while Chen and Wang\cite{chen2007statistical} proposed a method based on 2-D seriate images to discover statistics of a badminton match.\\
\indent Player detection and tracking methods for sports videos \cite{mentzelopoulos2013active, yan2014automatic} have been proposed in the past. Held et al.\cite{held2016learning} also propose a method for handling occlusions between players. Wang et al. \cite{wang2016classifying} used tracking data of players in the game of basketball to perform offensive playcall classification while Cervone et al. \cite{cervone2014pointwise} did point-wise predictions and discussed defensive metrics. Unlike these approaches, our method uses only visual cues and does not rely on special camera setup or additional sensors. There has also been some work in analysing sports without reliance on anything other than the broadcast video. Sukhwani et al.\cite{sukhwani2016frame} and Ghosh et al.\cite{ghosh2017smart} computed frame level annotations in tennis videos, however, where Sukhwani et al. used a dictionary learning method to co-cluster available textual descriptions, Ghosh et al.\cite{ghosh2017smart} used the scoreboard extraction approach to perform the index based analysis. \vspace{1mm}\\ 
\textbf{Real-time Sports Analysis: }Zhong et. al.\cite{zhong2004real, zhong2001structure} present a real-time framework for scene detection and structure analysis for tennis and baseball games using compressed-domain processing techniques. Further, they introduce another real-time framework to detect the syntactic structures that are at a level higher than shots for the game of tennis. Ekin et al.\cite{ekin2003generic} propose a single generic real-time algorithm to detect play-break events for multiple sports (football, tennis, basketball, and soccer). Their algorithm uses shot-based generic cinematic features to detect the events.\\
\indent Our work is inspired by the work of \cite{ghosh2018towards} that proposes an end-to-end framework to automatically annotate badminton broadcast videos. Their work identifies various understandable metrics that can be computed for analyzing badminton matches which in turn help qualitative understanding of badminton games. We build on top of~\cite{ghosh2018towards}, and extend the work further by (a) covering wider match situations/tournaments/games (b) adapting the solution to real-time deployment and (c) a successful field trial in {\bf PBL 2019} with commentators and broadcasters appreciating the utility.

\section{Dataset}
\begin{figure*}[!htp]
    \centering
    \begin{subfigure}{\textwidth}
        \includegraphics[width=.2\linewidth, height=.15\linewidth]{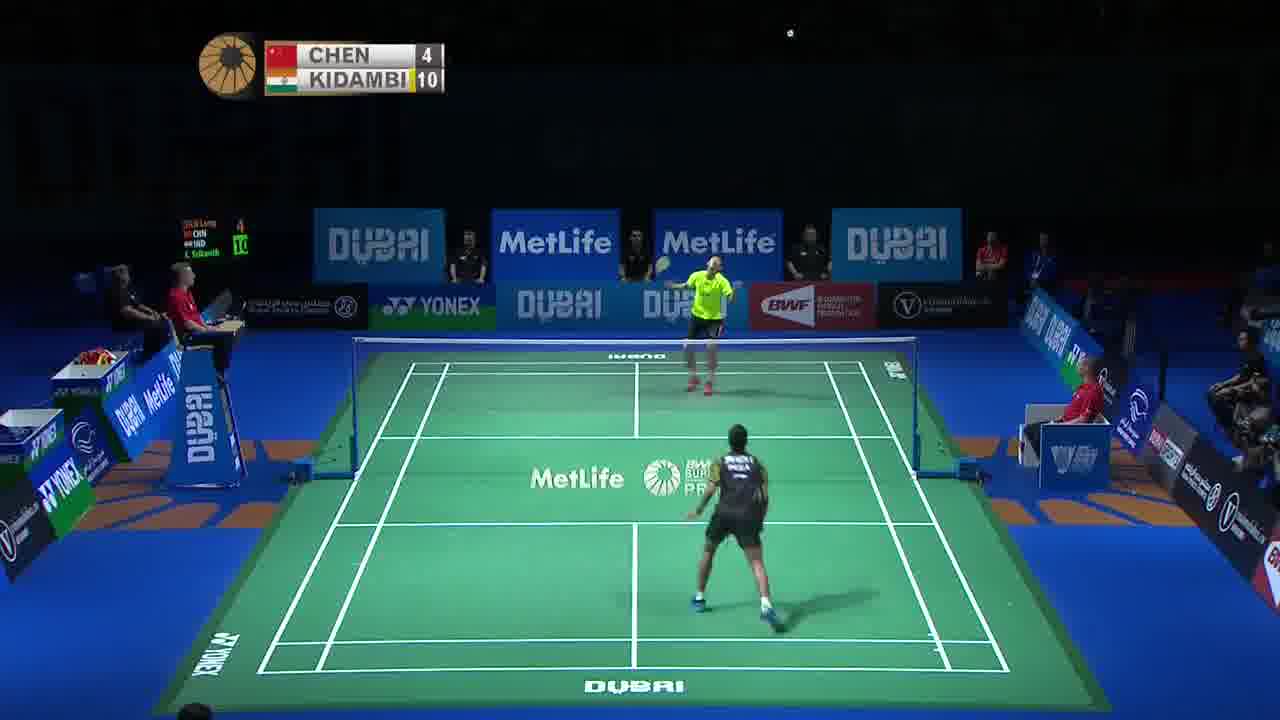}\hfill
        \includegraphics[width=.2\linewidth, height=.15\linewidth]{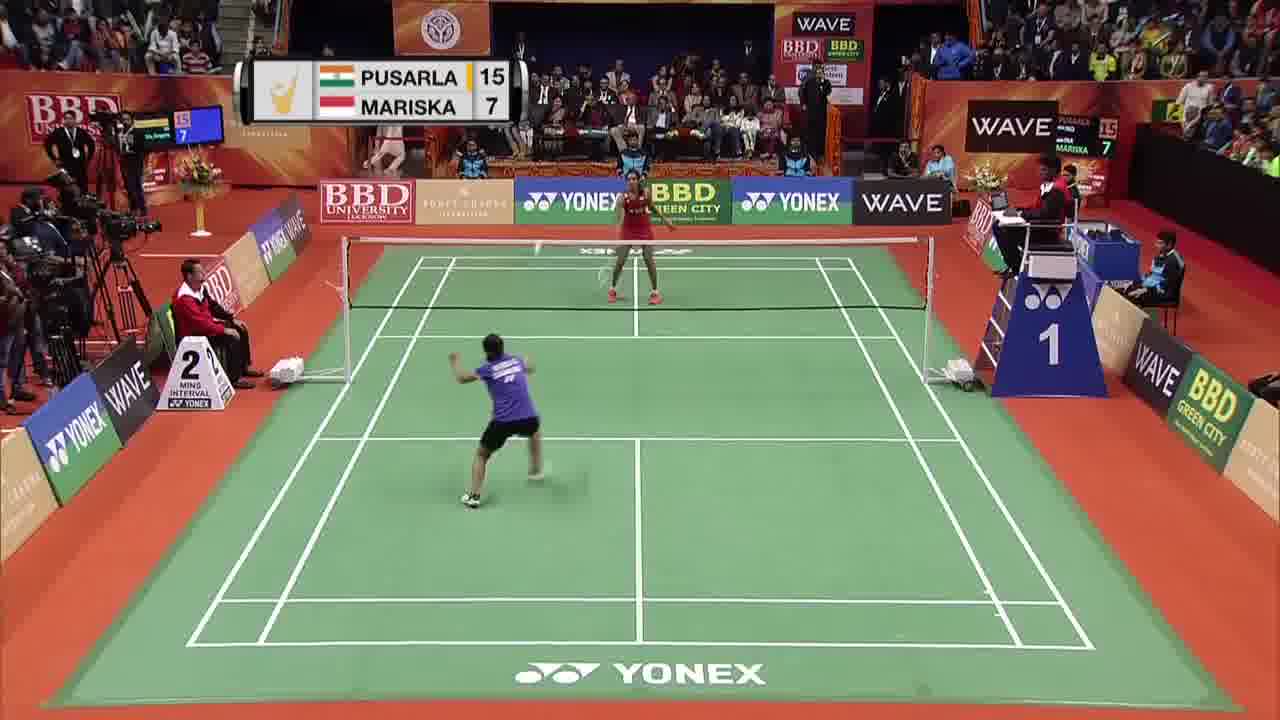}\hfill
        \includegraphics[width=.2\linewidth, height=.15\linewidth]{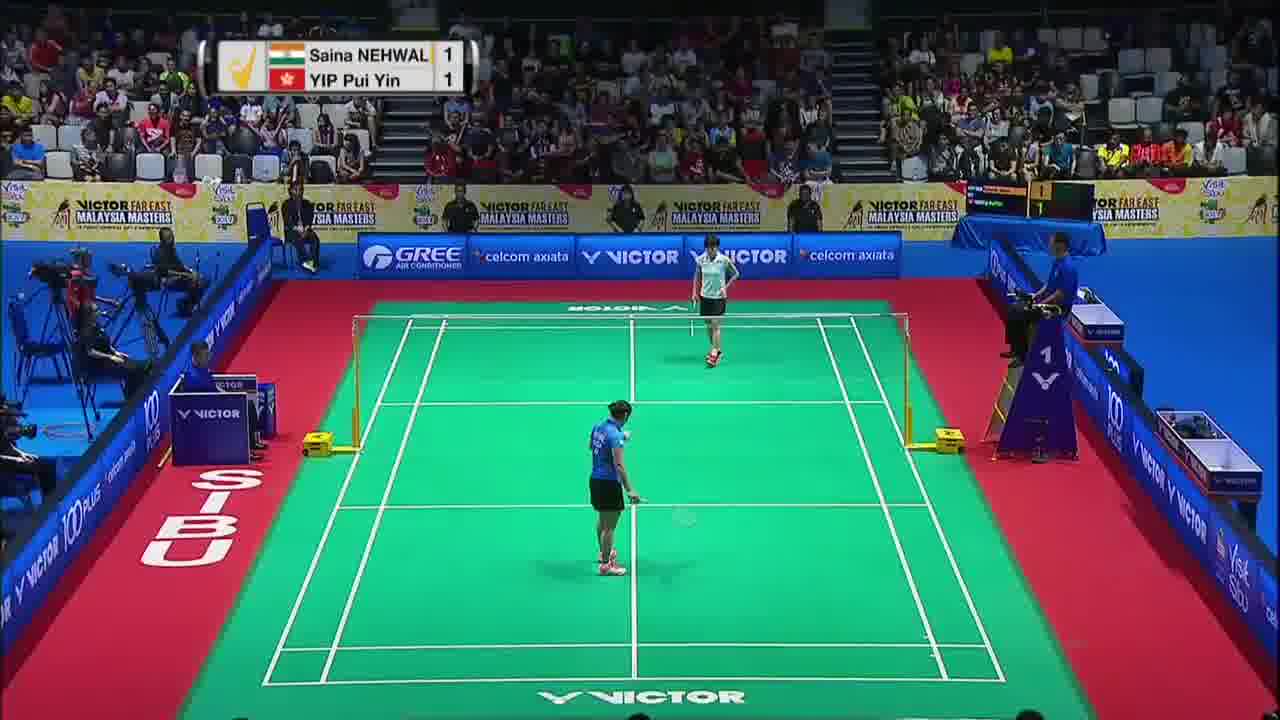}\hfill
        \includegraphics[width=.2\linewidth, height=.15\linewidth]{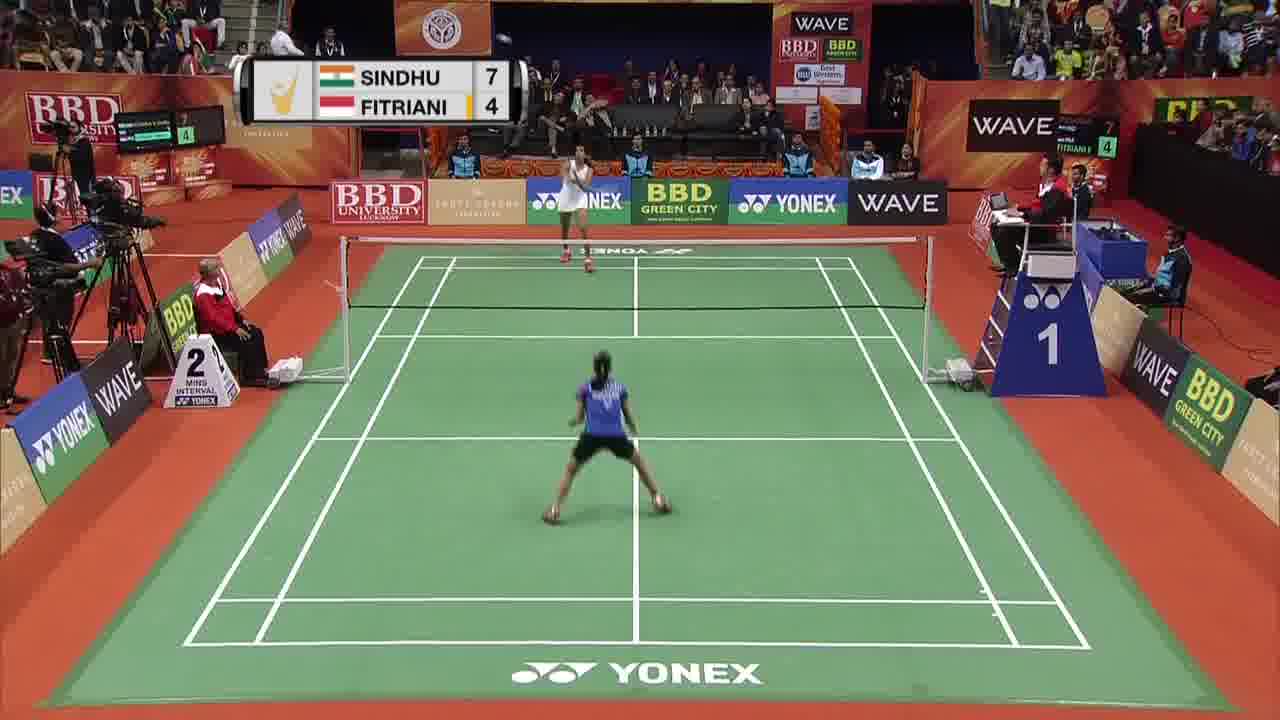}\hfill
        \includegraphics[width=.2\linewidth, height=.15\linewidth]{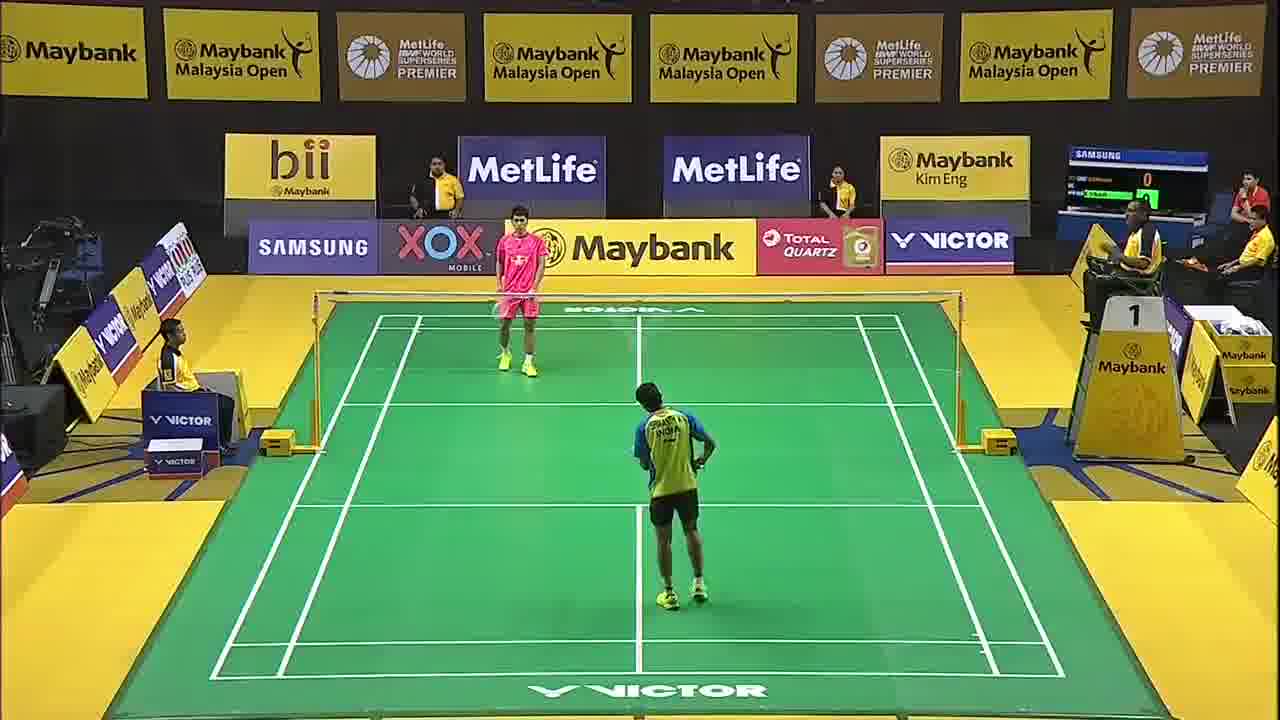}\hfill
    \end{subfigure}
    \begin{subfigure}{\textwidth}
        \includegraphics[width=.2\linewidth, height=.15\linewidth]{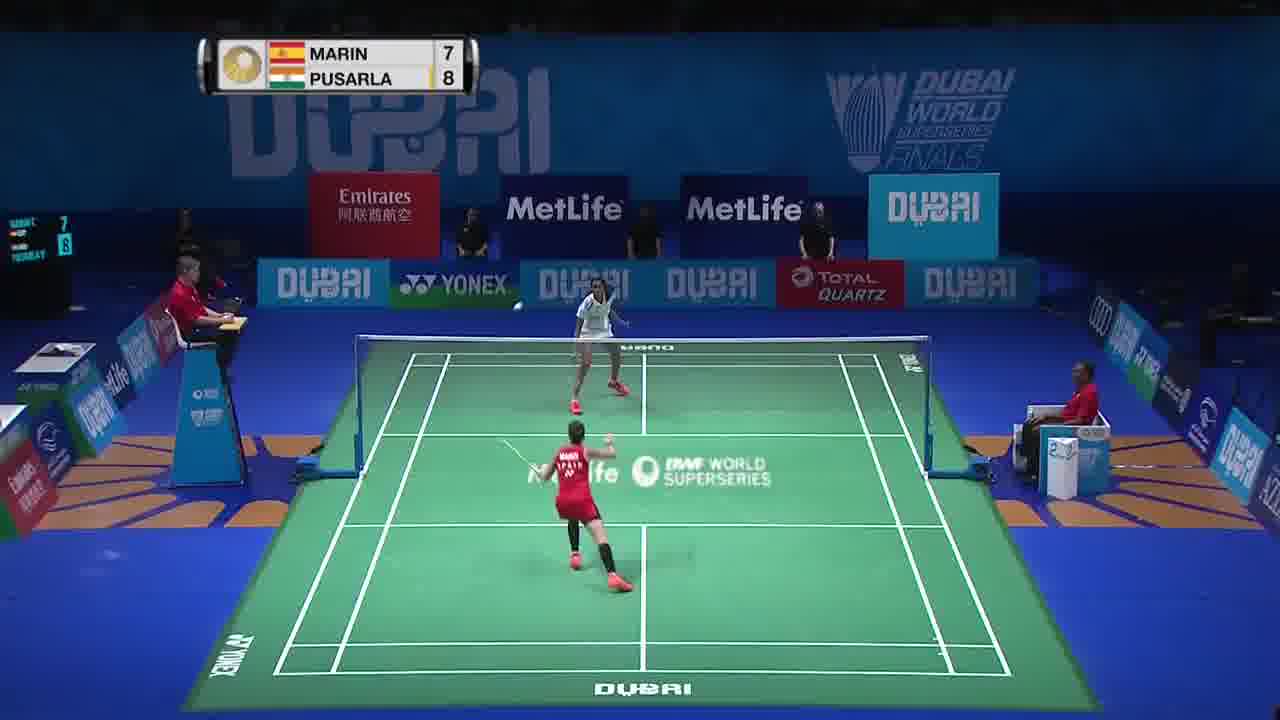}\hfill
        \includegraphics[width=.2\linewidth, height=.15\linewidth]{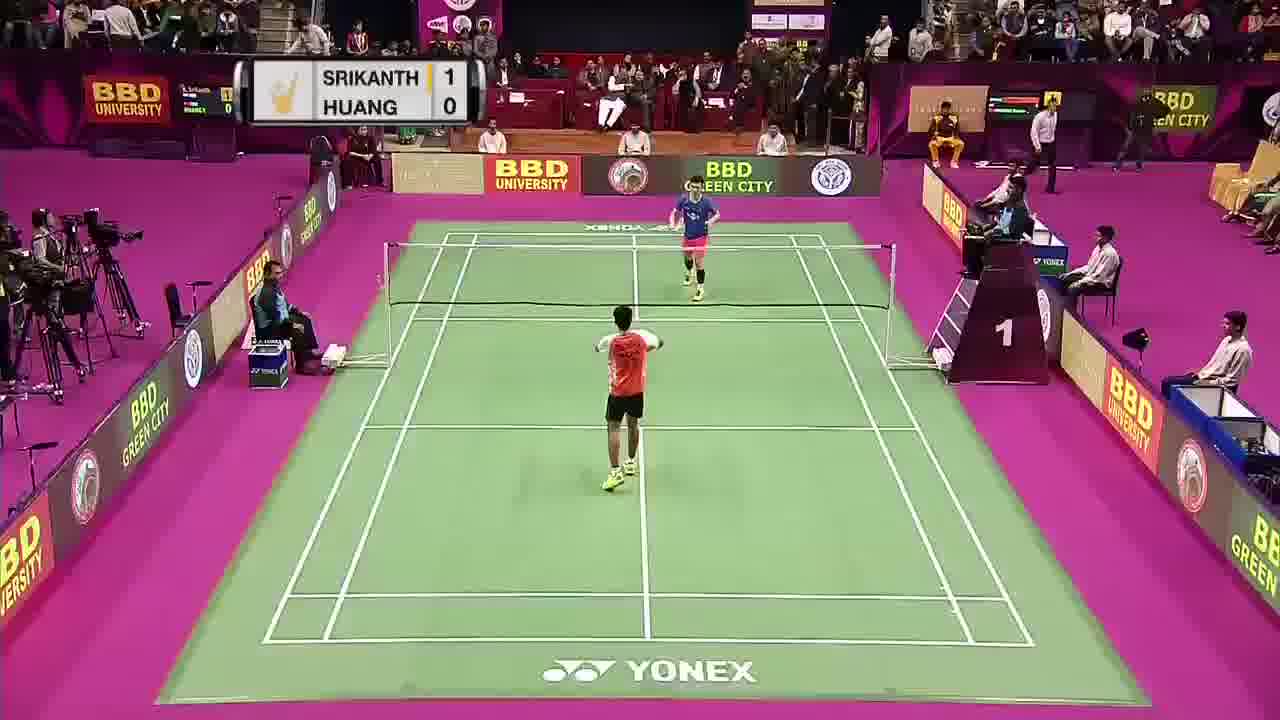}\hfill
        \includegraphics[width=.2\linewidth, height=.15\linewidth]{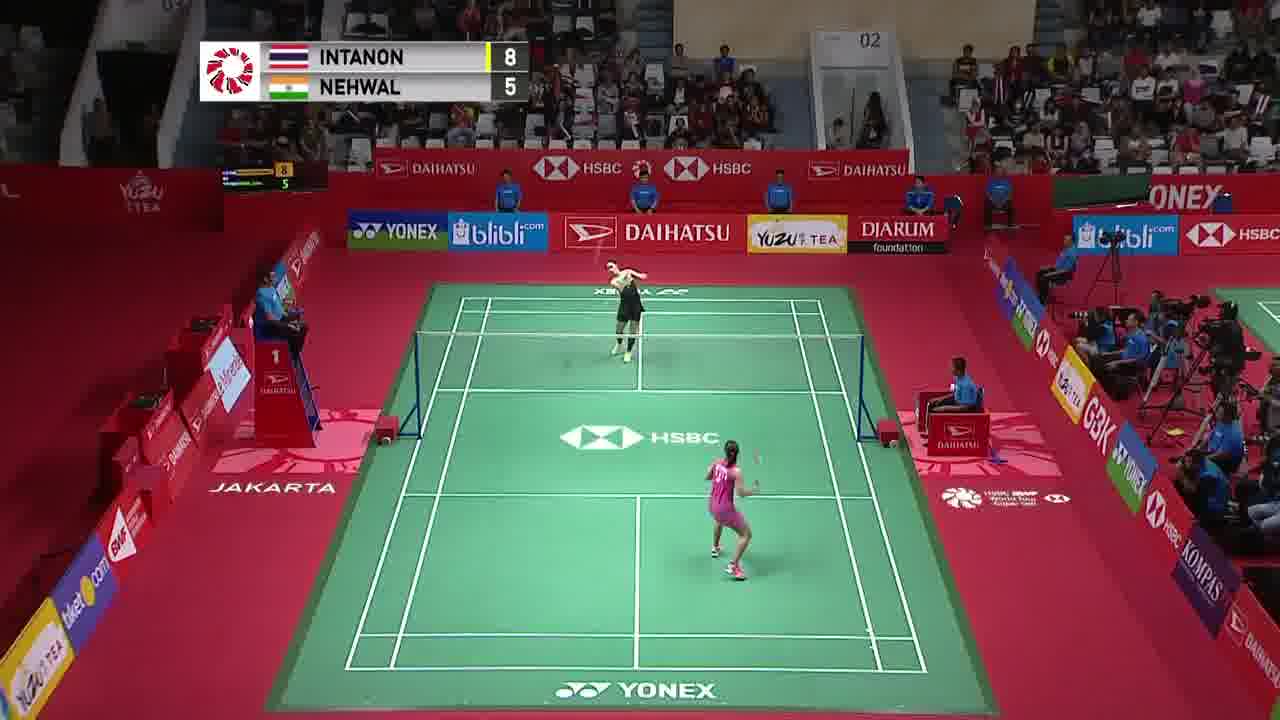}\hfill
        \includegraphics[width=.2\linewidth, height=.15\linewidth]{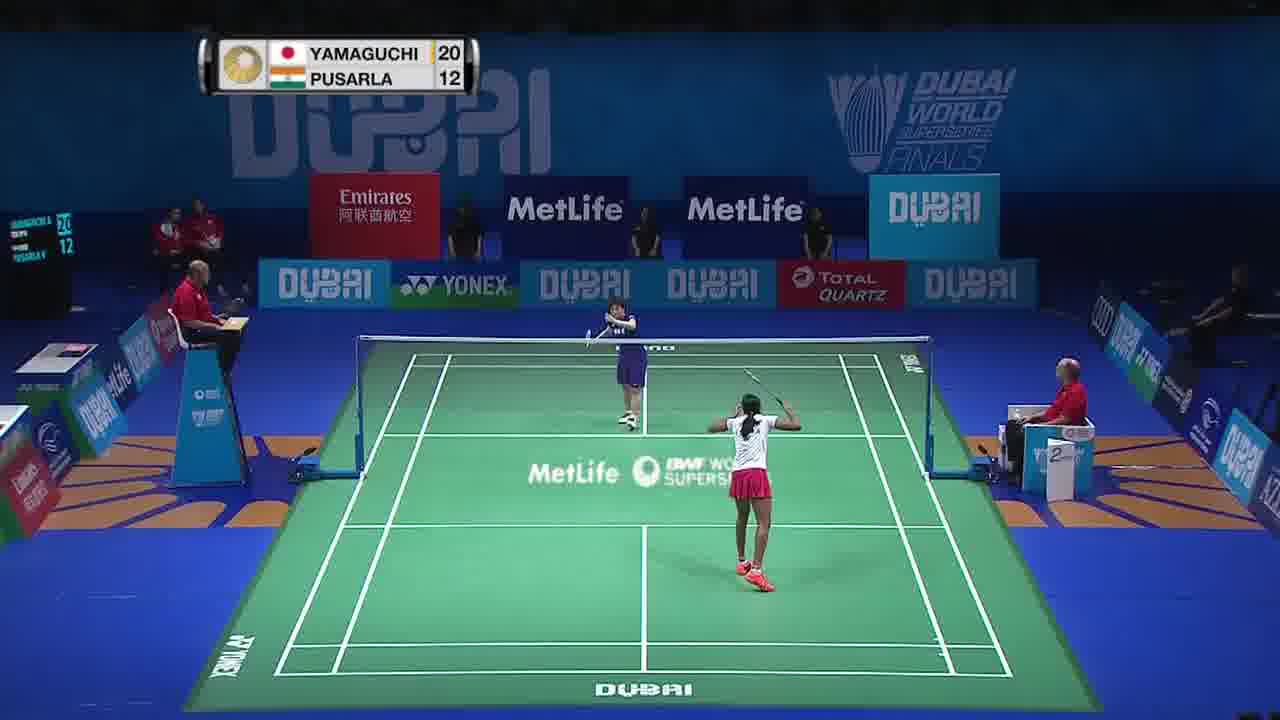}\hfill
        \includegraphics[width=.2\linewidth, height=.15\linewidth]{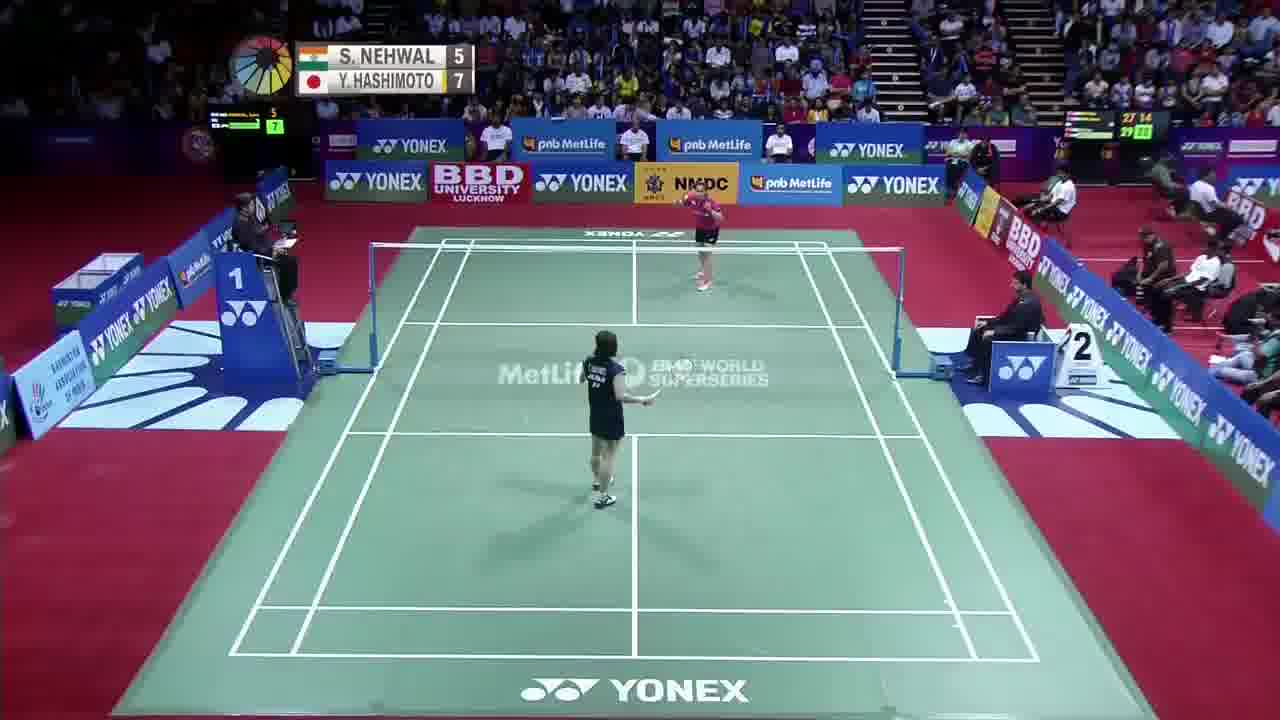}\hfill
    \end{subfigure}
    \caption{Our dataset contains various broadcast camera positions, illuminations settings, different backgrounds and court colors which helps generalize the framework for any kind of badminton match scenario. It should be observed that having audiences in the background may affect the player localization method if not generalized well.}\label{fig:datasetdetail}
    \vspace{-2mm}
\end{figure*}



A badminton game is divided into sets which could be either two or three in number. Each set has certain number of points or rallies. We work on a collection of 20 badminton match videos taken from the official Badminton World Federation (BWF) channel on YouTube\footnote{\href{https://www.youtube.com/user/bwf}{https://www.youtube.com/user/bwf}}. We focus on both ``singles" and ``doubles" (both men and women) matches played for two or three sets. These matches are typically around 60 to 90 minutes long depending upon the number of sets in the game. We also have our recorded badminton match video which consists of ground truth data for the distance covered by the players.
\par 
In the work by Ghosh et al. \cite{ghosh2018towards}, all the videos chosen to create the dataset were from the 2012 Summer Olympics. Due to this, all the match videos have the same court colors, illumination settings, and the same broadcasting camera angle. In order to introduce some variation, we create another dataset by taking videos from different tournaments held by the Badminton World Federation. For this, we select matches such that we try to cover all versatility of court structures, which includes different court colors, illumination settings, and all the variations in the broadcasting camera angle. All of these variations can be seen in fig \ref{fig:datasetdetail}. We choose the matches such that all opponents are unique, which ensures maximum variations in the dataset and also make sure that our analysis model does not overfit on a particular player. Table \ref{tab:dataset-comp} shows the comparison between our dataset and Ghosh et. al \cite{ghosh2018towards} dataset showing the differences between the matches.\par
To train and validate our approach, we took 10 matches each for both the singles and doubles matches, which contains 22 hours 16 minutes data, and annotate them for point segments and player bounding boxes. We split the ten matches into training set and testing set of 7 and 3 matches, respectively. We plan to release our dataset and annotations publicly post acceptance of the work.

\begin{table}
\resizebox{\columnwidth}{!}{%
\begin{tabular}{@{}l|ccc@{}}
\toprule
Component & \begin{tabular}[c]{@{}c@{}}Ghosh et. al.\cite{ghosh2018towards}\\ (Singles)\end{tabular} & \begin{tabular}[c]{@{}c@{}}Ours \\ (Singles)\end{tabular} & \begin{tabular}[c]{@{}c@{}}Ours\\ (Doubles)\end{tabular} \\ \midrule
Matches & 10 & 10 & 10 \\
Players & 20 & 13 & 20 \\
Camera Positions & 1 & 4 & 3 \\
Tournaments & 1 & 6 & 5 \\
Illumination Settings & 1 & 5 & 5 \\ \bottomrule
\end{tabular}%
}
\caption{A comparison of our datasets and the Badminton Olympic Dataset\cite{ghosh2018towards}. We show that our dataset is more diverse as we cover more tournaments, capturing multiple broadcast camera angles and a variety of illumination settings in the courts and background.}
\label{tab:dataset-comp}
\vspace{-5mm}
\end{table}

\subsection{Point Segmentation Annotation}
Broadcast badminton matches are divided into gameplays and replays. We define a ``rally" frame as a frame where the actual gameplay is happening and a ``non-rally" frame as a frame where the replay (highlights) has been shown, or players are resting, etc. (or in other words whether the gameplay is not happening). Generally, non-rally frames contain redundant information about the game and do not contribute much in the analysis part. Therefore, we first start classifying between rally and non-rally frames. We call this classification method as \textit{Point Segmentation}. This allows us to keep all the rally frames, group them into rallies according to the points scored by the players and discard the non-rally frames. To make the dataset usable for point segmentation, we annotate every frame of a match as a rally or non-rally frame using the ELAN video annotation tool \footnote{\href{https://tla.mpi.nl/tools/tla-tools/elan/}{https://tla.mpi.nl/tools/tla-tools/elan/}}. We annotate 23156 frames in total, where 15147 were non-rally frames and 8009 were rally frames. \par

\begin{figure*}[h]
    \centering
    \includegraphics[width=0.9\textwidth, height=0.4\textwidth]{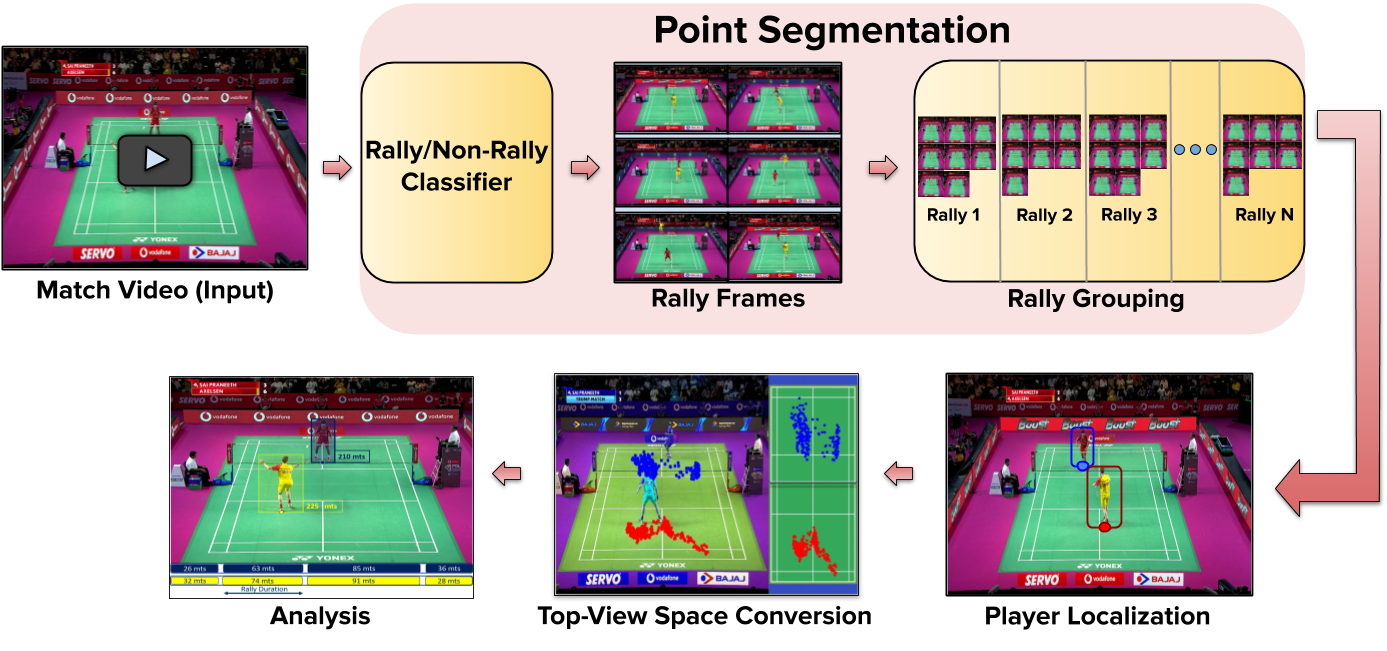}
    \caption{We propose a real-time end-to-end pipeline to analyze live braodcast match videos. Our pipeline consists of point segmentation, player localization, pictorial point summary and top-view space conversion to compute the distance and overall analysis.}
    \label{fig:approach}
    \vspace{-3mm}
\end{figure*}

\subsection{Player Bounding Boxes Annotation}
We annotate the dataset to get the bounding box for each player using the LabelImg\footnote{\href{https://github.com/tzutalin/labelImg}{https://github.com/tzutalin/labelImg}} graphical image annotation tool. We define the player's position in the court with respect to the broadcast camera point of view where one player plays from the near side, and other player plays from the far side of the camera, and we define them as bottom and top player respectively (see figure \ref{fig:intro} for more details). We manually annotate players' bounding boxes with two classes ``PlayerTop" and ``PlayerBottom" where ``PlayerTop" corresponds to the player on the far side of the court and while ``PlayerBottom" corresponds to the player on the near side of the court with respect to the viewpoint of the camera. We randomly pick 150 rally frames from each match and annotated them for player bounding boxes for both the players. Thus, we end up with 3000 (150 frames $\times$ 2 players $\times$ 10 matches) annotated frames in total. We use this annotation to learn player bounding boxes for the rest of the frames. We use the same annotation style for the doubles matches as well. As shown in figure \ref{fig:bbox}, occasionally, players are occluded or blurred due to the fast-paced nature of the game. Such occlusions are even more prevalent in the doubles matches where there are four players on court as opposed to the two players in a singles match. 

\subsection{Recorded Badminton Match Video}
Our analysis pipeline predicts distance covered by the players, and this forms the basis for other kinds of analysis (like average speed with which the players are moving). Therefore, it is imperative to rigorously evaluate the distance values predicted by our pipeline. Towards this end, we created a new dataset consisting of videos of amateur players playing badminton while wearing distance trackers. The distance trackers, which were accurate up to 1 meter, measured the distance covered by each player during each rally. We recorded 25 rallies and obtained the distance values for each player for all of them. All the videos were recorded at approximately the same broadcast angle as that of the singles and doubles datasets using a smartphone camera.

\section{Method}
As shown in figure \ref{fig:approach}, our pipeline is largely based on the approach proposed by Ghosh et al \cite{ghosh2018towards}. It comprises of 4 steps: (1) Point Segmentation (2) Player Localization (3) Top View Space Conversion and (4) Analysis. We optimize each of these steps independently to make them more robust while  simultaneously reducing latency. The improved pipeline can be used for real-time analysis of videos.
\subsection{Point Segmentation}
The very first step of our pipeline is to keep the rally frames and discard the non-rally frames (which are redundant for our analysis) from the given match video. Rally frames are usually always shot from the broadcast camera view whereas the non-rally frames can include close-ups of players, additional graphics (score cards) and unusual camera angles. These differences help distinguish these frames. Unlike Ghosh et al \cite{ghosh2018towards} who use HOG + SVM, we use a convolutional neural network based classifier to classify rally and non-rally frames. We finetune the ResNet-18\cite{he2016deep} model (pre-trained on ImageNet\cite{deng2009imagenet} dataset) by changing the last layer of this network to label the frame as rally frame or non-rally frame. As shown in table \ref{tab:comparison}, our CNN based method is both more accurate and also exhibits lower latency compared to HOG + SVM. 

At test time, we classify the frames retrieved from the match video and number the rally frames in a sequential manner. We assume that a rally is at least 3 seconds long and there is a minimum of 3 seconds gap between two rallies. This allows us to group frames of the match into individual rallies by separating sets of continuous rally frames with at least 3 seconds of non-rally frames in between them. Thus Point Segmentation gives us a set of rallies from the match. We number each rally sequentially as rally-1, rally-2 etc.

\subsection{Player Localization}
 Once we obtain individual rallies, we need to find player location in each frame of each rally. A YOLOv3\cite{yolov3} network is finetuned for these two classes. In order to overcome the constraint of only having 3000 annotated samples for player localization, we use transfer learning to train the YOLOv3 network. We first pre-train the YOLOv3 network on PASCAL-VOC dataset\cite{everingham2010pascal} which has 20 classes including the ``Person" class. Further, we finetune the YOLOv3 network on our own dataset to achieve better accuracy. We used the same method for both the single's and double's matches to get the location of the player. However, as can be seen in figure \ref{fig:bbox}, player localization is a very difficult problem to solve in doubles matches. \par
 
 
 \iftrue
\begin{figure}[!htp]
    \centering
    \begin{subfigure}{\linewidth}
        \includegraphics[width=.5\linewidth, height=.3\linewidth]{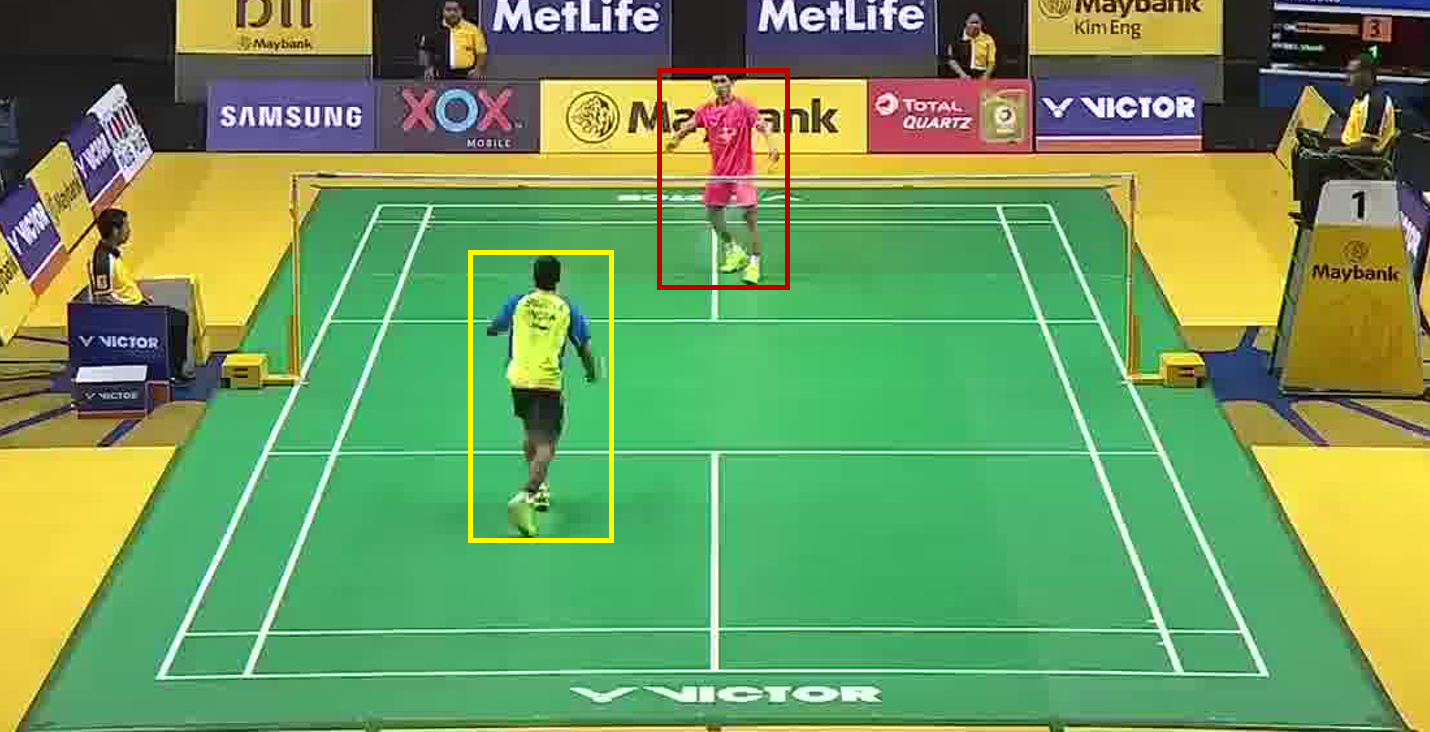}\hfill
        \includegraphics[width=.5\linewidth, height=.3\linewidth]{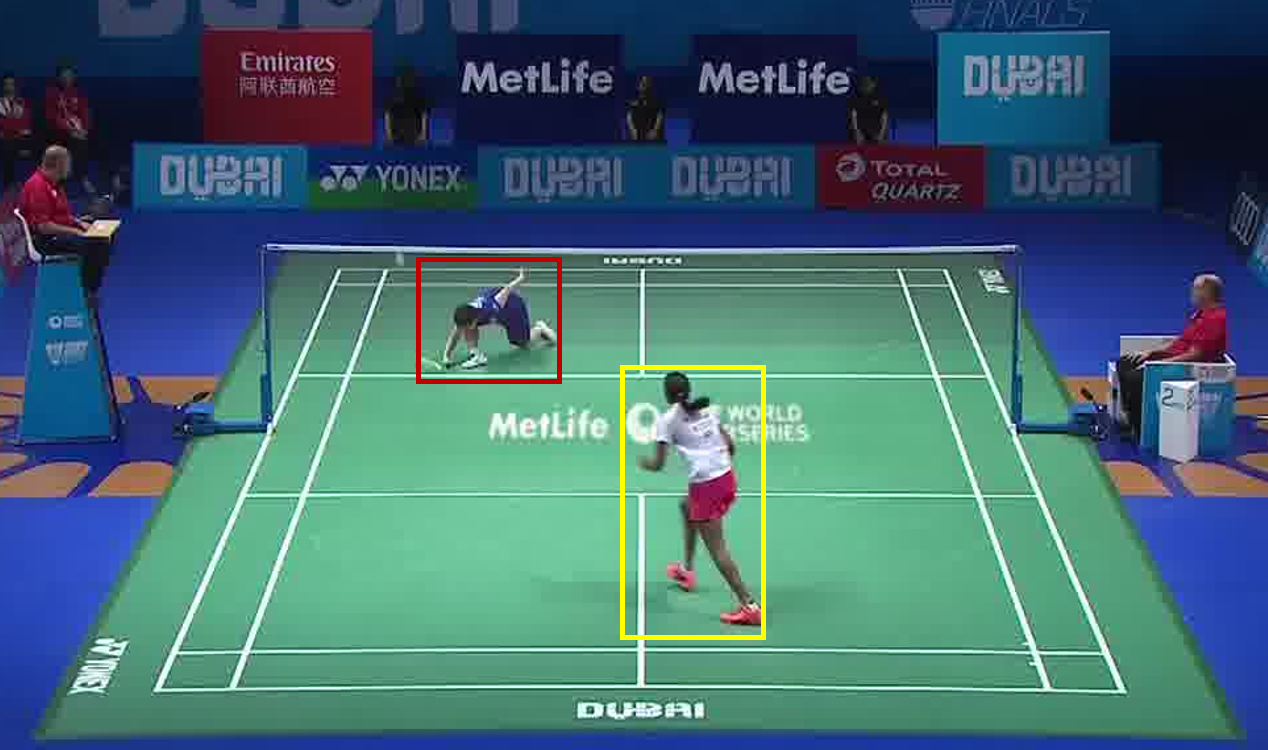}\hfill
    \end{subfigure}
    \begin{subfigure}{\linewidth}
        \includegraphics[width=.5\linewidth, height=.3\linewidth]{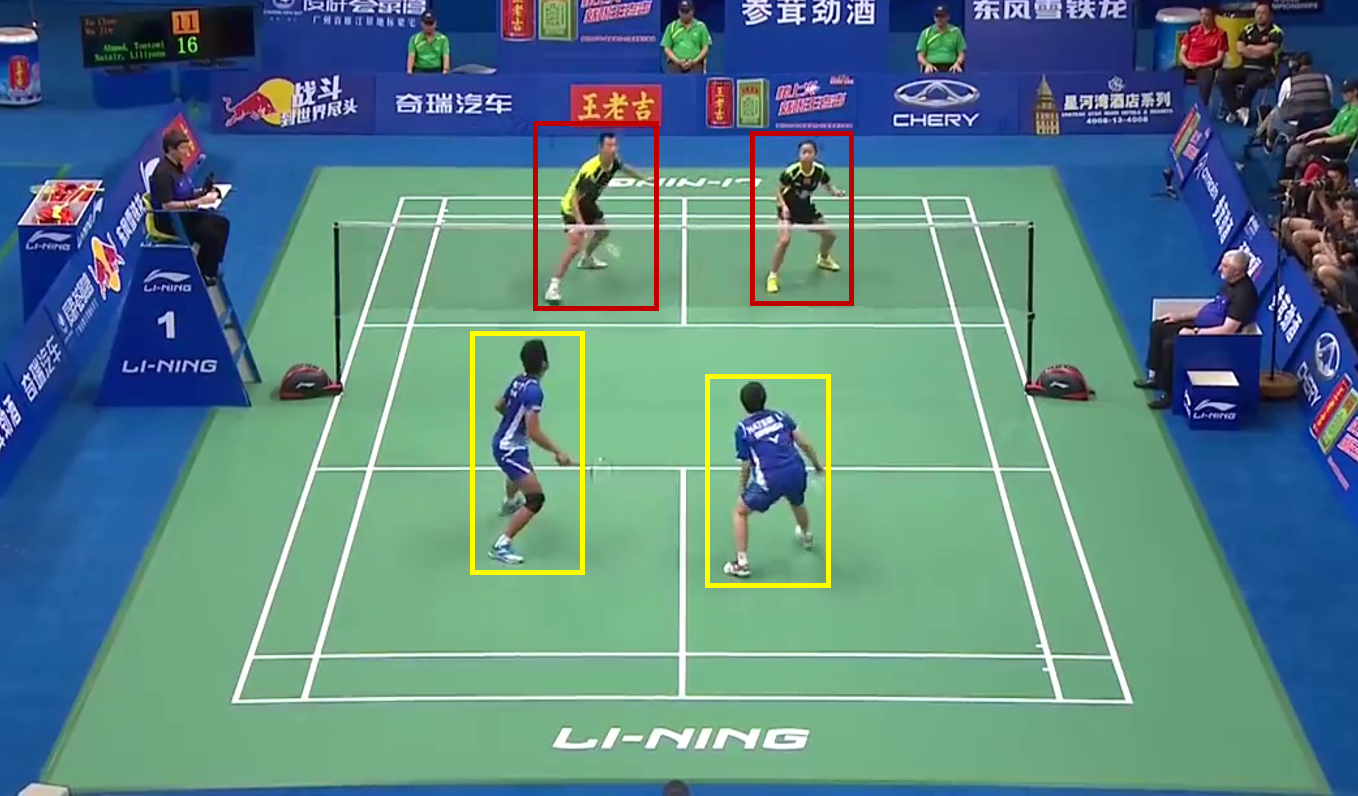}\hfill
        \includegraphics[width=.5\linewidth, height=.3\linewidth]{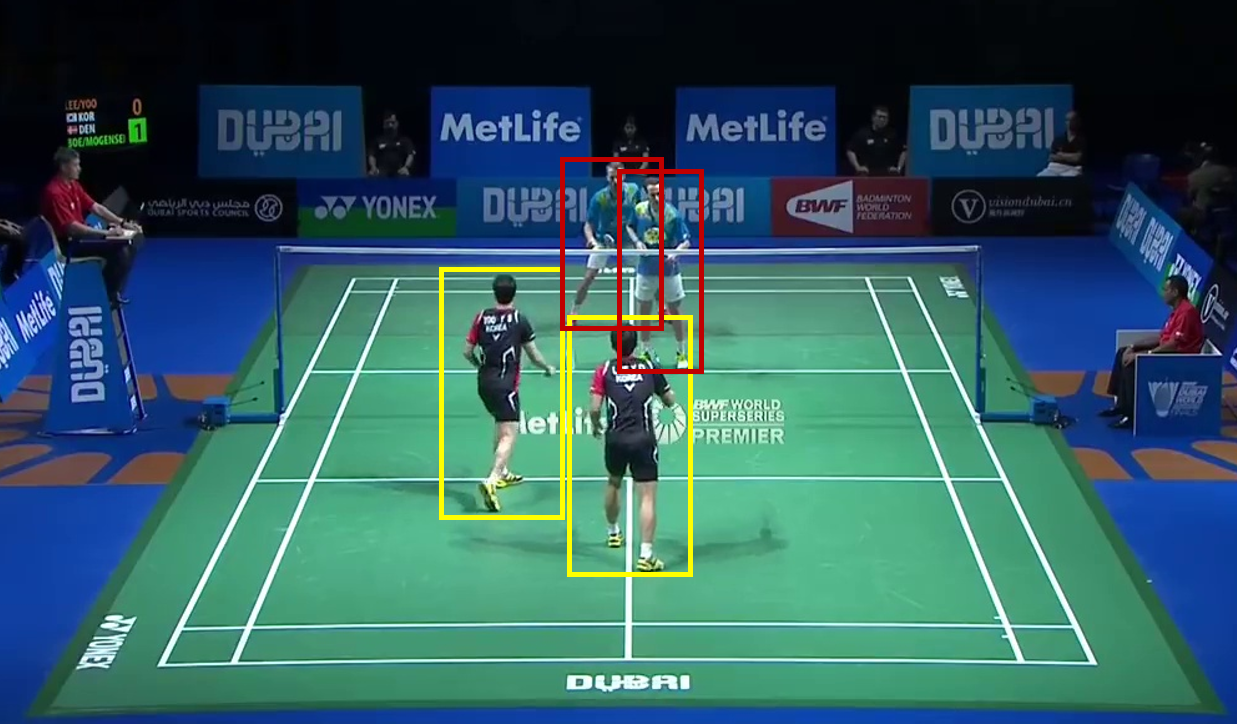}\hfill
    \end{subfigure}
    \caption{Output of player tracking method. It should be noted that this method works well, even when there are levels of occlusion and irrespective of player poses.}\label{fig:bbox}
\end{figure}
\fi

 Using a convolutional network for point segmentation and YOLOv3 for player localization helps us build a fast and robust pipeline that performs better than \cite{ghosh2018towards} both in terms of accuracy of predictions and latency. The improved accuracy as well as inference time make it possible to use our pipeline for a real-time implementation. Please refer to table \ref{tab:comparison} for more details.

\begin{table}[!htb]
    \begin{subtable}{\columnwidth}
    \sisetup{table-format=1.2}
        \resizebox{\textwidth}{!}{%
        \begin{tabular}{@{}c|cccc@{}}
        \toprule
        Method & F1-Score & Precision & Recall & \begin{tabular}[c]{@{}c@{}}Match Latency\\ (Minutes)\end{tabular} \\ \midrule
        HOG+SVM & 95.44 & 97.83 & 91.02 & 57.24 \\
        CNN & 97.23 & 98.06 & 92.34 & 16.75 \\ \bottomrule
        \end{tabular}%
        }
        \caption{Point Segmentation Method}
        \vspace{3mm}
    \end{subtable}
\smallskip
    \begin{subtable}{\columnwidth}
    \sisetup{table-format=1.2}
        \resizebox{\textwidth}{!}{%
        \begin{tabular}{@{}c|ccc@{}}
        \toprule
        Method & \begin{tabular}[c]{@{}c@{}}mAP@0.5\\ Player Top\end{tabular} & \begin{tabular}[c]{@{}c@{}}mAP@0.5\\ Player Bottom\end{tabular} & \begin{tabular}[c]{@{}c@{}}Rally Latency\\ (Seconds)\end{tabular} \\ \midrule
        Faster R-CNN & 96.90 & 97.85 & 76.63 \\
        YOLOv3 & 98.07 & 98.13 & 0.14 \\ \bottomrule
        \end{tabular}%
        }
        \caption{Player Bounding Box Method}
    \end{subtable}
    \caption{Comparison of methods between Ghosh et al.\cite{ghosh2018towards} and ours. We show that our method improves in terms of both latency and accuracy.}
    \label{tab:comparison}
    \vspace{-3mm}
\end{table}

\subsection{Top View Space}
We know that the displacement of both the players would manifest differently in the camera coordinates as one player is near side of the camera and another one is on the far side of the camera. In order to find out the distance or other kind of analysis, we need to switch the view of player positions from broadcast camera view to another view where the displacement of the player is not scaled by the near side or far side of the player position. We choose the top view of the badminton court where we map the player locations from broadcast camera view. So, using homography techniques, we map the 2-dimensional broadcast view camera coordinates to top view court coordinates. By doing this, we get an equivalent linear top view for the player locations. To perform homography calculation, we find the court lines from the video frame using Hough line finding method\cite{hough} and get the position of court corners. We compute a homography matrix between these four court corners and top view four court corners to get a mapping from camera coordinates to top view space coordinates. Finally, we map the player locations from broadcast camera view to top view. Please refer to figure \ref{fig:homography} for output details.

\iftrue
\begin{figure}[!htbp]
    \centering
    \begin{subfigure}{0.475\textwidth}
        \includegraphics[width=1\linewidth, height=.8\linewidth]{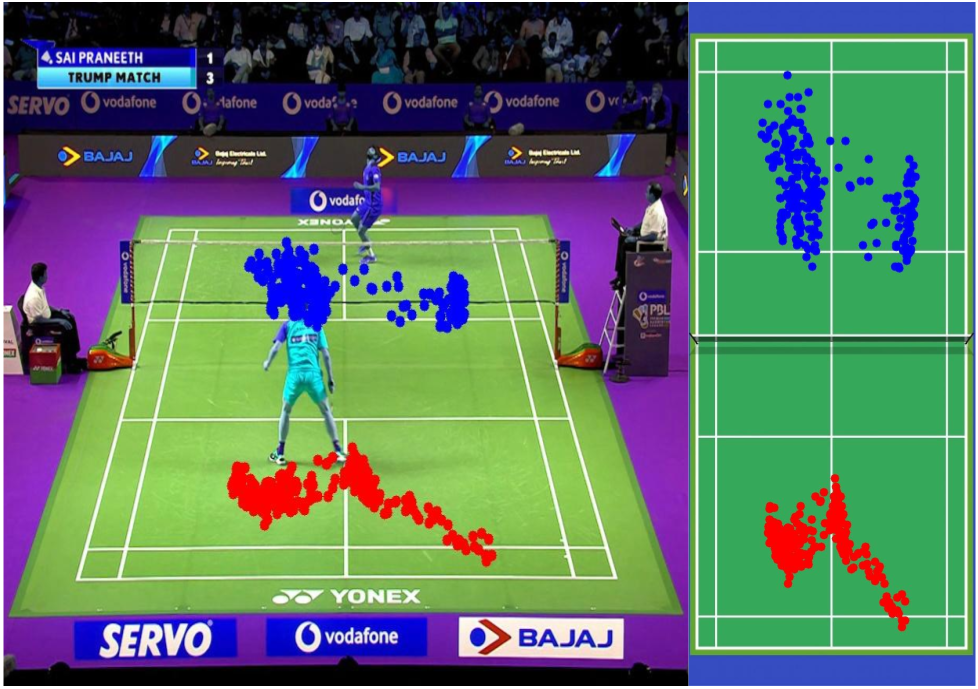}
    \end{subfigure}
    \caption{Conversion of player locations from camera coordinates to top view co-ordinates. The top-view space provides the player's position and footwork information around the court for a particular gameplay. This view can also be seen as heatmap of the players for that particular rally. \textit{(Best viewed in color)}}
    \label{fig:homography}
    \vspace{-5mm}
\end{figure}
\fi

\subsection{Analysis}

\noindent \textbf{Players Heatmap: } The output of top-view space conversion gives us insight into the footwork of the players around the court. This top-view can also be seen as a heatmap of the players' locations. This heatmap can be useful for the players and their coaches to analyze the game. \\

\noindent \textbf{Player Distance Calculation: } To compute the distance covered by the player in a particular rally, we utilize our player tracks in top view space. Now the player locations are in 2-dimensional space and we take the ``center bottom" point of the player bounding box as a proxy for the player's current location.

According to the standard badminton rules\footnote{\href{http://www.worldbadminton.com/rules/}{http://www.worldbadminton.com/rules/}}, the length and width of the badminton court is 13.41 meters and 6.1 meters respectively. From this information, we know that from the top left corner of the court to the top right corner of the court, the distance of pixels is 6.1 meters and similarly, from the top left corner of the court to the bottom left corner of the court the distance is 13.41 meters. We use this information to calculate the distance of the player covered in meters. We calculate the euclidean distance between the locations of the player in successive frames for both players. Now we have the pixel-based distance between two successive frames. We use unitary method to scale this pixel-based distance into meters to get the actual distance values. We add all the distances (meters) successively for the whole rally to calculate the distance for each player in that particular rally. The distance calculation of a rally for one player is formulated as: \[\sum_{i=1}^{n-1} \sqrt{(x_{i+1} - x_i)^2 \times \frac{6.1}{D_h} \ +\ (y_{i+1} - y_i)^2 \times \frac{13.41}{D_v}}\] where $n$ is the number of frames in the rally, $x_i$ and $y_i$ are the co-ordinates of the player in the top-view space at $i^{th}$ frame, $D_h$ is the pixel-based distance from top-left corner to the top-right corner (horizontal distance) and $D_v$ is the pixel-based distance from top-left corner to the bottom-left corner (vertical distance). For the game of double's, we have two players on each side of the court i.e. two ``PlayerTop" (pt) and two ``PlayerBottom" (pb). Therefore, for the first frame, we initialize $pt_1$ and $pt_2$ for the two top players and $pb_1$ and $pb_2$ for the two bottom players respectively. In subsequent frames, we use the player's previous position to track them individually. Again, we use the method same as singles to calculate the distance. \\

\noindent \textbf{Player Average Speed Calculation: } To calculate the average speed achieved by each of the players in a particular rally, we use the distance covered by the players. As we know the number of frames in a particular rally, we find out the total time elapsed (25 frames per second) and divide the player distance covered by the total time. 

\section{Evaluations}
We evaluate two different aspects of our proposed pipeline: accuracy of predictions and latency.
\subsection{Player Distance Calculation Evaluation}
To evaluate the accuracy of our proposed pipeline in terms of the distance covered by the players in a particular rally, we used our recorded badminton match video which has ground truth data of 25 rallies for the distance covered by the players. We use our proposed pipeline to obtain predicted distance values for each rally and compare them to the ground truth values. We used the same evaluation metric as \cite{edgecomb2006comparison}, which uses paired t-tests to calculate differences between the ground truth distances and predicted distances. We report the mean, standard deviation, minimum, and maximum error between the ground truth and predicted values for both ``Player Top" and ``Player Bottom". We also report the t-stats value and degree of freedom of the paired t-test calculation between the ground truth and predicted values. Please refer to table \ref{tab:freq} for the report.

\begin{table}[!htbp]
\resizebox{\columnwidth}{!}{%
\begin{tabular}{@{}c|cccccc@{}}
\toprule
Player Position & Mean & Std & Min & Max & t-stats & df \\ \midrule
Player Top & 4.5 & 1.18 & 2 & 6 & 0.20 & 24 \\
Player Bottom & 4.73 & 1.76 & 1 & 3 & -1.46 & 24 \\ \bottomrule
\end{tabular}%
}
\caption{We show the difference between the actual distance and the predicted distance covered by the players in our own recorded videos. It should be noted that maximum error we got is 6 meters and minimum error is 1 meter for player top and player bottom respectively.}
\label{tab:freq}
\vspace{-3mm}
\end{table}

\subsection{Real-time Analysis Evaluation}
\begin{figure}[t]
    \centering
    \includegraphics[width=0.50\textwidth, height=0.35\textwidth]{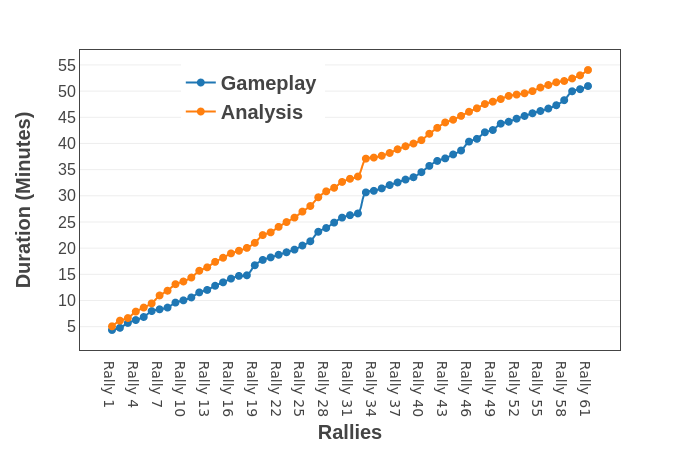}
    \caption{Comparison between time taken for gameplay and time taken for analysis in a simulated environment for a single pre-recorded game. The lower line denotes the time taken for gameplay and the upper line shows the time taken for analysis. We can see that the analysis curve closely follows the gameplay curve with a small time-lag. \textit{(Best viewed in color)}}
    \label{fig:timecomp}
\end{figure}

The proposed pipeline was successfully used in a real-world setting to analyze live broadcast matches in real-time during the Premier Badminton League 2019. As reproducing the exact same scenario is not feasible, we use a simulation of the real-time scenario to evaluate the performance (with respect to latency) of our model. Generally, as live matches are streamed, the camera recording the match captures 25 frames per second and thus the match video file size increases with time. In order to simulate the real-time scenario, we fed the pre-recorded match videos to our proposed pipeline at precisely 25 frames per second, sequentially. Thus our models did not have access to all the frames at the same time. We ran this simulation on 61 rallies (totaling up to 55 minutes) and timed our models. Figure \ref{fig:timecomp} shows the results of this simulation. 

\section{Experiments}

We use the live feed of the broadcast video match from Premier Badminton League - 2019 (courtesy of Star Sports India\footnote{\href{http://www.pbl-india.com/}{http://www.pbl-india.com/}}). We used frames from the live video in batches of 20 for our experiments. We attempt to extract the analysis from the data we got using the methods mentioned above. We have segmented the frames into either rally or non-rally frame using point segmentation method which in the end, gives us rally directories as soon as the rally is over in the live match, where each directory contains rally frames for that specific rally. Then we pass the rally frames to the player localization method to get the player bounding boxes. We perform homography to change the space from broadcast camera view to top-view. We perform each analysis on frame level for the accessed file.

On an average, one rally is 10 to 12 seconds long and it contains 250 to 300 rally frames. For a real time implementation, it is not feasible to process these many frames in one go. Hence, to solve this problem for this scenario, we process every $3^{rd}$ rally frame for player localization method. As there are 25 frames per second in the broadcast match video, we assume that there is no significant change in player location if we take every third rally frame. After getting the rally directories from the above point segmentation method we pass the rally directories to the player localization method to get the player location for each of the frames in a parallel manner. Using this rally-level parallelism for analysis helped us to achieve the goal in the real-time scenario. 

\section{Field Trial} We now show the result of our analysis which is computed in real-time at Premier Badminton League - 2019 and was aired on ``Hotstar", which is the official Star sports channel. We computed and showed the distances covered by the players in each set (typically a group of 15 - 30 rallies). Please refer to the figure \ref{fig:result} for output details.

\iftrue
\begin{figure}[!htbp]
    \centering
    \begin{subfigure}{0.475\textwidth}
        \includegraphics[width=1\linewidth, height=.6\textwidth]{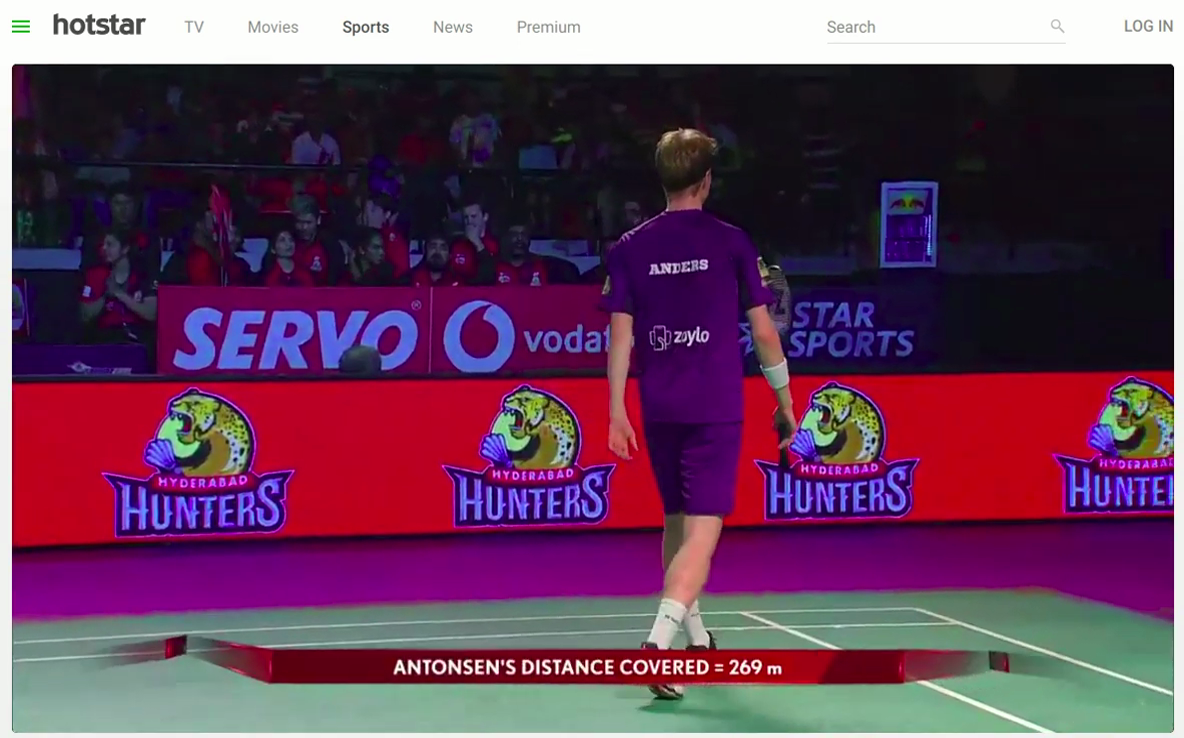}
    \end{subfigure}
    \begin{subfigure}{0.475\textwidth}
        \includegraphics[width=1\linewidth, height=.6\textwidth]{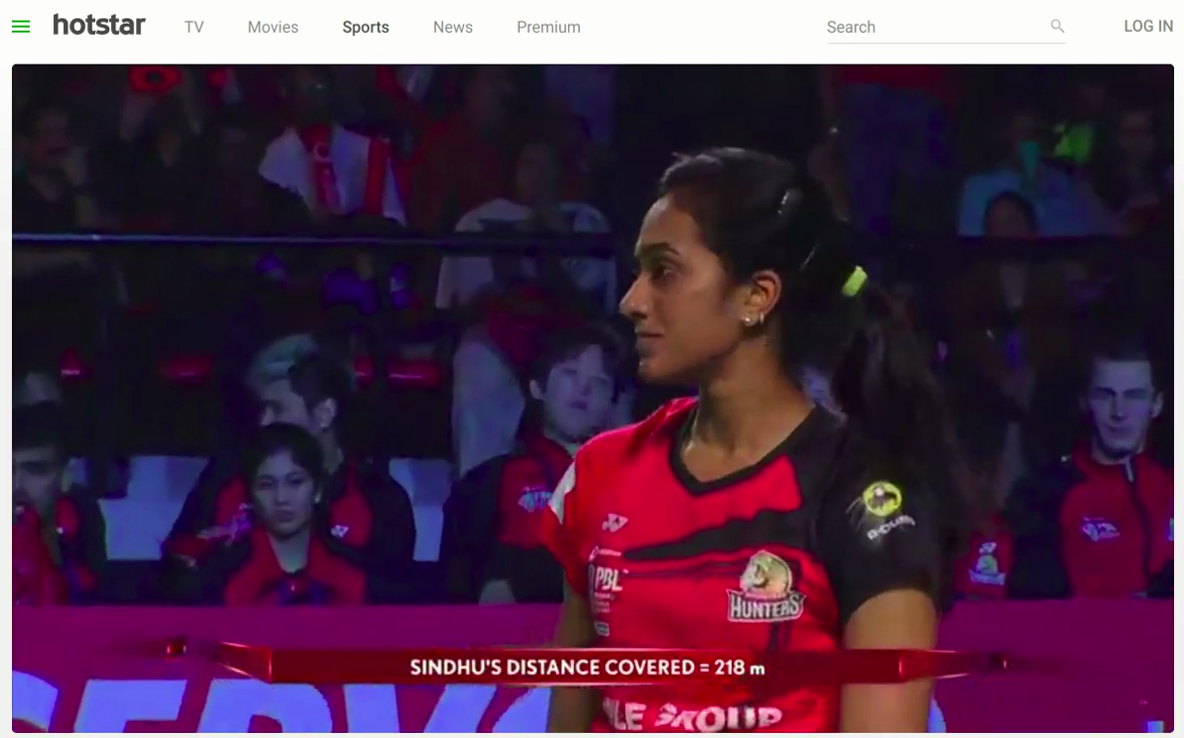}
    \end{subfigure}
    \caption{Distance covered by the players as computed by our pipeline in real-time at PBL-2019. The distance covered by the player is shown in metres. These images are taken from the Hotstar application during one of the games of PBL 2019.}
    \label{fig:result}
\end{figure}
\fi

\section{Conclusion and Future Work}
All existing work in badminton video analysis suffers from the drawback that it relies on recorded match videos and cannot be used to display the analysis or the computed match statistics in real time. In this work, we present an efficient and robust pipeline to analyze badminton matches (both singles and doubles) in real-time. We perform 3 different types of analysis: (1) heatmap generation (2) distance covered by players and (3) average speed of player movement. We optimize each component of our pipeline to ensure that it maintains high accuracy while also minimizing inference latency. We show the evaluations for the player distance calculation and its implementation in real-time scenario. We also empirically show that processing every third frame of the video is sufficient to perform the desired analysis. The real-time implementation of these kinds of analyses not only enhances the viewer experience but also helps players improve their game.

Another problem that hinders research in the area of sports analysis, and more specifically, badminton analysis, is the lack of labelled datasets. We introduce 3 different datasets to help combat this problem. Two of our datasets (singles and doubles) consist of Badminton World Federation match videos and feature professional badminton players and official courts and recording equipment. Both these datasets are annotated for Point Segmentation and Player Localization. The third dataset consists of amateur players recorded by using a smartphone camera from the approximately same broadcast angle as the other datasets. This dataset also consists of ground truth data for the distance covered by the players in each rally (which was obtained from the distance trackers worn by the players during gameplay). We hope that access to these datasets helps spur further research into real-time analysis of badminton videos. 

The proposed pipeline was validated in a real-world setting when it was successfully used to analyze live broadcast matches in real-time during the Premier Badminton League 2019 ({\bf PBL 2019}). In the future, we would like to extend this work to other types of badminton analysis like shot classification, shot recommendation etc. and build a pipeline that can do them in real-time. We are also optimistic about extending this pipeline to other racquet sports like tennis and squash.

{\small
\bibliographystyle{./IEEEtran}
\bibliography{IEEEfull}

\begin{thebibliography}{10}
\providecommand{\url}[1]{#1}
\csname url@samestyle\endcsname
\providecommand{\newblock}{\relax}
\providecommand{\bibinfo}[2]{#2}
\providecommand{\BIBentrySTDinterwordspacing}{\spaceskip=0pt\relax}
\providecommand{\BIBentryALTinterwordstretchfactor}{4}
\providecommand{\BIBentryALTinterwordspacing}{\spaceskip=\fontdimen2\font plus
\BIBentryALTinterwordstretchfactor\fontdimen3\font minus
  \fontdimen4\font\relax}
\providecommand{\BIBforeignlanguage}[2]{{%
\expandafter\ifx\csname l@#1\endcsname\relax
\typeout{** WARNING: IEEEtran.bst: No hyphenation pattern has been}%
\typeout{** loaded for the language `#1'. Using the pattern for}%
\typeout{** the default language instead.}%
\else
\language=\csname l@#1\endcsname
\fi
#2}}
\providecommand{\BIBdecl}{\relax}
\BIBdecl

\bibitem{tjondronegoro2004highlights}
D.~Tjondronegoro, Y.-P.~P. Chen, and B.~Pham, ``Highlights for more complete
  sports video summarization,'' \emph{IEEE Multimedia}, 2004.

\bibitem{giancola2018soccernet}
S.~Giancola, M.~Amine, T.~Dghaily, and B.~Ghanem, ``Soccernet: A scalable
  dataset for action spotting in soccer videos,'' in \emph{CVPR Workshops},
  2018.

\bibitem{maksai2016players}
A.~Maksai, X.~Wang, and P.~Fua, ``What players do with the ball: A physically
  constrained interaction modeling,'' in \emph{CVPR}, 2016.

\bibitem{ghosh2018towards}
A.~Ghosh, S.~Singh, and C.~Jawahar, ``Towards structured analysis of broadcast
  badminton videos,'' in \emph{WACV}.\hskip 1em plus 0.5em minus 0.4em\relax
  IEEE, 2018.

\bibitem{felsen2017will}
P.~Felsen, P.~Agrawal, and J.~Malik, ``What will happen next? forecasting
  player moves in sports videos,'' in \emph{ICCV}, 2017.

\bibitem{sharma2017automatic}
R.~A. Sharma, V.~Gandhi, V.~Chari, and C.~Jawahar, ``Automatic analysis of
  broadcast football videos using contextual priors,'' \emph{SIVP}, 2017.

\bibitem{gandhi2010real}
H.~Gandhi, M.~Collins, M.~Chuang, and P.~Narasimhan, ``Real-time tracking of
  game assets in american football for automated camera selection and motion
  capture,'' \emph{Procedia Engineering}, 2010.

\bibitem{shitrit2011tracking}
H.~B. Shitrit, J.~Berclaz, F.~Fleuret, and P.~Fua, ``Tracking multiple people
  under global appearance constraints,'' in \emph{ICCV}.\hskip 1em plus 0.5em
  minus 0.4em\relax IEEE, 2011.

\bibitem{ghanem2012context}
B.~Ghanem, M.~Kreidieh, M.~Farra, and T.~Zhang, ``Context-aware learning for
  automatic sports highlight recognition,'' in \emph{ICPR}.\hskip 1em plus
  0.5em minus 0.4em\relax IEEE, 2012.

\bibitem{yoshikawa2010automated}
F.~Yoshikawa, T.~Kobayashi, K.~Watanabe, and N.~Otsu, ``Automated service scene
  detection for badminton game analysis using chlac and mra,'' \emph{World
  Academy of Science, Engineering and Technology}, 2010.

\bibitem{chu2017badminton}
W.-T. Chu and S.~Situmeang, ``Badminton video analysis based on spatiotemporal
  and stroke features,'' in \emph{ICMR}.\hskip 1em plus 0.5em minus 0.4em\relax
  ACM, 2017.

\bibitem{mlakar2017analyzing}
M.~Mlakar and M.~Lu{\v{s}}trek, ``Analyzing tennis game through sensor data
  with machine learning and multi-objective optimization,'' in
  \emph{Proceedings of the 2017 ACM International Joint Conference on Pervasive
  and Ubiquitous Computing and Proceedings of the 2017 ACM International
  Symposium on Wearable Computers}.\hskip 1em plus 0.5em minus 0.4em\relax ACM,
  2017.

\bibitem{chen2007statistical}
B.~Chen and Z.~Wang, ``A statistical method for analysis of technical data of a
  badminton match based on 2-d seriate images,'' \emph{Tsinghua science and
  technology}, 2007.

\bibitem{mentzelopoulos2013active}
M.~Mentzelopoulos, A.~Psarrou, A.~Angelopoulou, and
  J.~Garc{\'\i}a-Rodr{\'\i}guez, ``Active foreground region extraction and
  tracking for sports video annotation,'' \emph{Neural processing letters},
  2013.

\bibitem{yan2014automatic}
F.~Yan, J.~Kittler, D.~Windridge, W.~Christmas, K.~Mikolajczyk, S.~Cox, and
  Q.~Huang, ``Automatic annotation of tennis games: An integration of audio,
  vision, and learning,'' \emph{Image and Vision Computing}, 2014.

\bibitem{held2016learning}
D.~Held, S.~Thrun, and S.~Savarese, ``Learning to track at 100 fps with deep
  regression networks,'' in \emph{ECCV}.\hskip 1em plus 0.5em minus 0.4em\relax
  Springer, 2016.

\bibitem{wang2016classifying}
K.-C. Wang and R.~Zemel, ``Classifying nba offensive plays using neural
  networks,'' in \emph{Proceedings of MIT Sloan Sports Analytics Conference},
  2016.

\bibitem{cervone2014pointwise}
D.~Cervone, A.~D’Amour, L.~Bornn, and K.~Goldsberry, ``Pointwise: Predicting
  points and valuing decisions in real time with nba optical tracking data,''
  in \emph{Proceedings of the 8th MIT Sloan Sports Analytics Conference,
  Boston, MA, USA}, 2014.

\bibitem{sukhwani2016frame}
M.~Sukhwani and C.~Jawahar, ``Frame level annotations for tennis videos,'' in
  \emph{ICPR}.\hskip 1em plus 0.5em minus 0.4em\relax IEEE, 2016.

\bibitem{ghosh2017smart}
A.~Ghosh and C.~Jawahar, ``Smarttennistv: Automatic indexing of tennis
  videos,'' in \emph{NCVPRIPG}.\hskip 1em plus 0.5em minus 0.4em\relax
  Springer, 2017.

\bibitem{zhong2004real}
D.~Zhong and S.-F. Chang, ``Real-time view recognition and event detection for
  sports video,'' \emph{Journal of Visual Communication and Image
  Representation}, 2004.

\bibitem{zhong2001structure}
------, ``Structure analysis of sports video using domain models,'' in
  \emph{ICME}.\hskip 1em plus 0.5em minus 0.4em\relax Citeseer, 2001.

\bibitem{ekin2003generic}
A.~Ekin and M.~Tekalp, ``Generic play-break event detection for summarization
  and hierarchical sports video analysis,'' in \emph{ICME}, 2003.

\bibitem{he2016deep}
K.~He, X.~Zhang, S.~Ren, and J.~Sun, ``Deep residual learning for image
  recognition,'' in \emph{CVPR}, 2016.

\bibitem{deng2009imagenet}
J.~Deng, W.~Dong, R.~Socher, L.-J. Li, K.~Li, and L.~Fei-Fei, ``Imagenet: A
  large-scale hierarchical image database,'' in \emph{CVPR}, 2009.

\bibitem{yolov3}
J.~Redmon and A.~Farhadi, ``Yolov3: An incremental improvement,'' \emph{arXiv},
  2018.

\bibitem{everingham2010pascal}
M.~Everingham, L.~Van~Gool, C.~K. Williams, J.~Winn, and A.~Zisserman, ``The
  pascal visual object classes ({VOC}) challenge,'' \emph{IJCV}, 2010.

\bibitem{hough}
R.~O. Duda and P.~E. Hart, ``Use of the hough transformation to detect lines
  and curves in pictures,'' Sri International Menlo Park Ca Artificial
  Intelligence Center, Tech. Rep., 1971.

\bibitem{edgecomb2006comparison}
S.~Edgecomb and K.~Norton, ``Comparison of global positioning and
  computer-based tracking systems for measuring player movement distance during
  australian football,'' \emph{Journal of science and Medicine in Sport}, 2006.

\end{thebibliography}
}




\end{document}